\documentclass[preprint]{article}

\usepackage{neurips_2023}

\usepackage[utf8]{inputenc} % allow utf-8 input
\usepackage[T1]{fontenc}    % use 8-bit T1 fonts
\usepackage{hyperref}       % hyperlinks
\usepackage{url}            % simple URL typesetting
\usepackage{booktabs}       % professional-quality tables
\usepackage{amsfonts}       % blackboard math symbols
\usepackage{nicefrac}       % compact symbols for 1/2, etc.
\usepackage{microtype}      % microtypography
\usepackage{xcolor}         % colors
\usepackage{listings}
\usepackage{cite}
\usepackage{graphicx}
\usepackage{hyperref}

\hypersetup{
    pdftitle={Large Malaysian Language Model Based on Mistral for Enhanced Local Language Understanding},
    pdfauthor={Husein Zolkepli, Aisyah Razak, Kamarul Adha, Ariff Nazhan},
    pdfsubject={Natural Language Processing, Large Language Models, Machine Learning},
    pdfkeywords={language models, malay natural language processing, deep learning},
    colorlinks=true,
    linkcolor=blue,
    citecolor=blue,
    urlcolor=blue
}
\title{Large Malaysian Language Model Based on Mistral for Enhanced Local Language Understanding}

\author{
  Husein Zolkepli\thanks{husein@mesolitica.com} \and 
  Aisyah Razak\thanks{aisyahrazak171@gmail.com} \and
  Kamarul Adha\thanks{kamarul.adha360@gmail.com} \and
  Ariff Nazhan\thanks{ariffnzhn@gmail.com}
}

\begin{document}

\maketitle

\begin{abstract}
  In this paper, we present significant advancements in the pretraining of Mistral 7B, a large-scale language model, using a dataset of 32.6 GB, equivalent to 1.1 billion tokens. We explore the impact of extending the context length, releasing models with context lengths of 4096 and 32768 tokens, and further refining performance with a specialized 16384 context length instruction-tuned model, we called it Malaysian Mistral.

  Our experiments demonstrate the efficacy of continue pretraining and the influence of extended context lengths on Mistral 7B's language understanding capabilities. Additionally, we release a model specifically tuned with a 16384 context length instruction, showcasing its potential for capturing nuanced language intricacies.

  Furthermore, our research contributes to the benchmarking of Malaysian Mistral against prominent language models, including ChatGPT3.5 and Claude 2. We present compelling results indicating Malaysian Mistral's superior performance on Tatabahasa (Malay grammar) test set, particularly when fine-tuned with instructions.

  All models released at \href{https://huggingface.co/collections/mesolitica/malaysian-mistral-7b-6528f2ec825f4bba46c1700c}{HuggingFace Mesolitica Malaysian Mistral 7B Collection}.
\end{abstract}

\section{Introduction}
The evolution of artificial intelligence (AI) has witnessed transformative breakthroughs, from the introduction of "Attention is All You Need" \cite{vaswani2023attention} with the Transformer architecture, to subsequent advancements like GPT-2, and the revolutionary ChatGPT. These models have sparked immense interest and curiosity in the AI landscape, pushing the boundaries of natural language understanding and generation.

In response to this dynamic landscape, Mistral AI emerged, unveiling its initial model, Mistral 7B \cite{jiang2023mistral}. Notably, Mistral 7B showcased superior performance, surpassing benchmarks set by Llama 2 13B across various tasks and even outperforming Llama 1 34B on numerous benchmarks. Impressively, it approached the performance of CodeLlama 7B on code-related tasks while maintaining proficiency in English language tasks. However, an identified gap in its capabilities was the limited understanding of Malaysian context.

\begin{itemize}
  \item \textbf{Fine-tuning Mistral 7B:} Utilizing the computational power of 8x A100 GPUs on a Standard\_ND96amsr\_A100\_v4 Azure instance, we conducted extensive fine-tuning on Mistral 7B. The process involved training the model using context lengths of 4096 and 32768 on a substantial 32.6 GB Malaysian context dataset.

  \item \textbf{Multi-turn Instruction-Tuned Model:} Crafting local context multiturn chat dataset using ChatGPT3.5 and ChatGPT4, we employed Neural Machine Translation to translate the dataset. This approach enhances Malaysian Mistral's proficiency in multi-turn conversations, contributing to its adaptability across a wide range of local context tasks and coding.
\end{itemize}

\section{Related Work}

\subsection{English-Centric Bias in Large Language Models}

The majority of open-source Large Language Models (LLMs) exhibit a significant bias towards the English language, with minimal representation and training on Malay datasets. An analysis of the widely utilized Common Crawl dataset reveals a mere 0.0742\% contribution from the Malay language based on \href{https://commoncrawl.github.io/cc-crawl-statistics/plots/languages}{CC-MAIN-2023-50} index. This English-centric bias poses a substantial challenge for applications requiring robust language understanding in Malay, prompting the need for dedicated research and development in this domain.

\subsection{Existing Malay Language Models}

While the Malay natural language processing (NLP) landscape lacks a dedicated Large Language Model, notable efforts have been made by Mesolitica in the development of specific Malay language models. Notable among these are the Malay Causal Language Model, Malay T5, and Malay Masked Language Model. These models, while contributing significantly to the Malay NLP toolkit \cite{Malaya}, are distinct from comprehensive Large Language Models and have limitations in capturing extensive context and nuances.

\subsection{Absence of a Malay Large Language Model}

Despite the existence of specialized models for Malay, a notable gap remains in the absence of a dedicated Malay Large Language Model. The current state of affairs hinders the progress of research and applications requiring a deeper understanding of the Malay language. A comprehensive Large Language Model for Malay is essential to bridge this gap, enabling advancements in various natural language processing tasks and fostering the inclusive representation of Malay in the AI landscape.

\section{Pre-Training Procedure}

\subsection{Public Data}

\subsubsection{Wikipedia}\label{sec:wikipedia}

We initiated the process by downloading the Malay Wikipedia dump from \url{https://dumps.wikimedia.org/}. This dump serves as a valuable resource for capturing diverse linguistic contexts in the Malay language.

The pertinent information from the Malay Wikipedia dump was extracted using the \url{https://github.com/attardi/wikiextractor} tool. This tool will parsed XML files into JSON line files and removed unnecessary XML tags.

We also obtained the English Wikipedia dataset from \url{https://huggingface.co/datasets/wikipedia}. Subsequently, we filtered documents containing keywords 'malay' or 'malaysia'. This targeted filtering ensures that the English dataset retains content relevant for the Malaysian context.

\subsubsection{Malay Language study articles}

To enrich our dataset, we incorporated the Malay dictionary, specifically the "Kamus Dewan Edisi Keempat." This authoritative source provides a comprehensive collection of Malay words, enhancing the language model's understanding of vocabulary and linguistic nuances.

We also included articles from \href{https://jendeladbp.my/}{JendelaDBP}. The selected articles cover various domains, including language (bahasa), literature (sastera), society (masyarakat), culture (budaya), economy (ekonomi), and Islamic studies (islam). This inclusion ensures that the language model is exposed to a broad spectrum of content, promoting a more holistic understanding of the Malay language within different contexts.

\subsubsection{Malaysia Government public documents}

For a comprehensive understanding of legislative proceedings, we incorporated data from the Malaysia Hansard. This authoritative source provides official transcripts of parliamentary debates, contributing to the language model's exposure to formal and legal discourse.

Legal documents from \url{https://lom.agc.gov.my} were included to provide the language model with insights into legal terminology, regulations, and official language usage within the Malaysian legal context.

To enhance the language model's grasp of political discourse, we utilized parliamentary records from \url{https://parlimen.gov.my}. This inclusion ensures exposure to discussions on national policies, legislation, and political matters.

To incorporate real-world data and statistics, we accessed datasets from \url{https://data.gov.my}. This enriches the language model's training data with diverse information on various aspects of Malaysian governance and public services.

Articles from the Mufti Wilayah contribute to the language model's understanding of religious discourse, ethical discussions, and Islamic perspectives on various topics.

To capture a broad range of government-related documents, we performed a systematic Google search for public PDFs from the 'gov.my' and 'edu.my' domains. This ensures inclusion of publicly available official documents, reports, and publications.

\subsubsection{Malaysia online articles}

To augment our dataset with a diverse range of Malaysian context, we employed web scraping techniques to gather public articles covering various topics. This comprehensive approach includes content related to gaming, education, blogging, politics, economy, sports, and a multitude of other subjects. The goal is to provide the language model with a broad and nuanced understanding of Malaysian perspectives, opinions, and discussions across a spectrum of domains. Complete list at \href{https://github.com/malaysia-ai/dedup-text-dataset/blob/main/mistral/combine-mistral.ipynb}{link here}.

\subsection{Deduplicating Data}

To minimize redundancy within our dataset, we implemented the MinHash algorithm utilizing the implementation available at \url{https://github.com/ChenghaoMou/text-dedup}.

We configured the MinHash algorithm with the following parameters:

\begin{table}[h]
  \centering
  \begin{tabular}{lccl}
    \hline
    \textbf{Parameter} & \textbf{Value} \\
    \hline
    num\_perm          & 256            \\
    threshold          & 0.95           \\
    hash\_func         & sha1           \\
    hash\_bits         & 64             \\
    \hline
  \end{tabular}
\end{table}

Complete deduplicating data implementation at \href{https://github.com/malaysia-ai/dedup-text-dataset?tab=readme-ov-file#text-dedup}{here}. All deduped dataset published at \href{https://huggingface.co/datasets/malaysia-ai/dedup-text-dataset}{malaysia-ai/dedup-text-dataset}.

\subsection{Postprocessing Data}

After the initial data collection and deduplication steps, we applied postprocessing techniques to refine the dataset for optimal training. The following steps were implemented:

\begin{itemize}
  \item Removal of Texts with HTTP Errors.

  \item Filtering Texts by Length, we removed texts less than 3 characters.

  \item Normalization of Whitespace, we replaced 6 spaces or more with 6 spaces.

  \item Normalization of Punctuation, we replaced 6 dots or more with 6 dots.
\end{itemize}

Complete postprocessing data implementation at \href{https://github.com/malaysia-ai/dedup-text-dataset?tab=readme-ov-file#postprocessing}{here}.

% https://github.com/malaysia-ai/dedup-text-dataset?tab=readme-ov-file#postprocessing

\subsection{Pre-Training phase}

\subsubsection{Tokenizing Data}

We adopted a packing technique for tokenization, wherein the dataset was segmented into sequences with context lengths of 4096 and 32768. The sequences were separated by the End-of-Sequence (EOS) token, indicating the end of one segment and the beginning of the next.

Complete tokenizing data implementation at \href{https://github.com/malaysia-ai/dedup-text-dataset/tree/main/mistral}{here}.

\subsubsection{4096 context length pre-training}\label{sec:4096-pretrained}

In our pre-training, we chose the causal language model approach. This method trains the model to predict the next word in a sequence from the previous context. The
objective during this phase is to maximize the likelihood of the entire sequence,

\[
  P(x_1, x_2, \ldots, x_T) = \prod_{t=1}^{T} P(x_t | x_1, x_2, \ldots, x_{t-1})
\]

We utilized Standard\_ND96asr\_v4 Azure instance which contains 8x A100 80 GB GPUs (NVLink 3.0). The pre-training hyperparameters are detailed below:

\begin{table}[h]
  \centering
  \begin{tabular}{lccl}
    \hline
    \textbf{Hyperparameter} & \textbf{Value} \\
    \hline
    DeepSpeed               & ZeRO-3 Offload \\
    Batch Size              & 20             \\
    Learning Rate           & constant 2e-5  \\
    Precision               & bfloat16       \\
    \hline
  \end{tabular}
\end{table}

Complete pre-training 4096 context length implementation at \href{https://github.com/mesolitica/malaya/tree/5.1/session/mistral#7b-4096-context-length}{here}.

\subsubsection{32768 context length pre-training}\label{sec:32768-pretrained}

For 32768 context length, we only trained on 10\% of the total dataset and use latest checkpoint from 4096 context length pretrained model from section ~\ref{sec:4096-pretrained}. We use the same Standard\_ND96asr\_v4 Azure instance which contains 8x A100 80 GB GPUs (NVLink 3.0). The pre-training hyperparameters are detailed below:

\begin{table}[h]
  \centering
  \begin{tabular}{lccl}
    \hline
    \textbf{Hyperparameter} & \textbf{Value} \\
    \hline
    DeepSpeed               & ZeRO-3 Offload \\
    Batch Size              & 3              \\
    Learning Rate           & constant 2e-5  \\
    Precision               & bfloat16       \\
    \hline
  \end{tabular}
\end{table}

Complete pre-training 32768 context length implementation at \href{https://github.com/mesolitica/malaya/tree/5.1/session/mistral#7b-32768-context-length}{here}.

% https://github.com/mesolitica/malaya/tree/5.1/session/mistral#7b-32768-context-length

\section{Supervised Instruction Fine-tuning Procedure}

\subsection{Generating Instruction Dataset}\label{sec:generating-instruction-dataset}

In our synthetic instruction dataset creation process, we employed ChatGPT3.5 and ChatGPT4 to generate diverse sets of instructions tailored to specific domains. ChatGPT3.5 was utilized to generate instructions with a focus on Malaysian context, contributing to a dataset enriched with localized language nuances. Additionally, ChatGPT3.5 was employed to generate pseudo Retrieval-Augmented Generation Multi-Turn instructions on Malaysian context, further diversifying the content pool.

For coding instructions and synthetic CommonQA, we leveraged the capabilities of ChatGPT4, utilizing its advanced language generation capabilities to create instructions that specifically pertain to coding tasks. This inclusion broadened the dataset's scope to encompass both language-related and technical instructions.

A notable aspect of this generation process was the tendency of these models to produce instructions in Indonesian. To address this, we implemented Neural Machine Translation, a powerful tool that seamlessly translated the instructions from Indonesian to standard Malay. This translation system demonstrated proficiency not only in translating between Indonesian and standard Malay but also in handling other languages such as English, Javanese, and Banjarese. Notably, it exhibited the ability to maintain the structural integrity of the original text, selectively translating only necessary components, such as programming code snippets, while preserving the overall text structure. We open-sourced the neural machine translation model at \href{https://huggingface.co/mesolitica/translation-t5-small-standard-bahasa-cased-code}{mesolitica/translation-t5-small-standard-bahasa-cased-code}.

\subsubsection{Coding Dataset}

We improved our coding dataset with original code samples from \url{https://huggingface.co/datasets/theblackcat102/evol-codealpaca-v1}, inspired by the WizardCoder \cite{luo2023wizardcoder}. To make the dataset more inclusive, we used ChatGPT4 to translate English coding questions into Malay and generate corresponding Malay answers. This approach expands the language model's capabilities, allowing it to understand and respond to coding queries in both English and Malay. Below is the prompt to translate the dataset into Malay and provide answers in Malay,

\begin{lstlisting}[breaklines=true]
  instruction = 'Please amend the subsequent Python script so that it includes ...'
  ins = chatgpt4(f'terjemah ke bahasa melayu: {instruction}')
  answer = chatgp4(ins + ', jawab dalam bahasa melayu')
\end{lstlisting}

While this process exhibited proficiency, there were instances where ChatGPT4 generated responses in Indonesian. To ensure linguistic consistency, we use our own Neural Machine Translation mentioned at ~\ref{sec:generating-instruction-dataset}, seamlessly translating the Indonesian responses to Malay.

Output from ChatGPT4,

\begin{lstlisting}[breaklines=true]
  Untuk mengubah skrip ini, kita Akkan menggunakan konsep pengulangan 'while' dan memastikan indeks tidak melebihi panjang senarai. Selain itu, kita harus memastikan skrip kita mengendalikan senarai kosong dengan betul.

  Berikut adalah cara untuk mengubahnya:

  ```python
    # Menetapkan senarai integer
    arr = [1, 2, 3, 4]

    # Tentukan panjang senarai
    n = len(arr)

    # Pastikan senarai tidak kosong
    if n > 0:
      i = 0
      # Melalui senarai dan keluarkan setiap elemen individu
      while i < n:
          print(arr[i])
          i += 1
    else:
      print("Senarai adalah kosong.")
  ```

  Dalam skrip ini, sebelum kita mula mengulangi, kita memeriksa sama ada `n`, yang merupakan panjang senarai, lebih besar daripada 0. Jika tidak (yang bermaksud senarai kosong), kita mencetak mesej yang sesuai.

  Kemudian, kita menggunakan pengulangan 'while' untuk mengulangi semasa indeks `i` kurang daripada panjang `n` senarai. Semasa setiap iterasi, kita mencetak elemen senarai dan menambahkan 1 ke indeks, supaya kita boleh bergerak ke elemen seterusnya pada iterasi seterusnya.

  Ini akan mengelakkan ralat indeks, kerana kita tidak akan pernah mencuba mengakses indeks yang melebihi had senarai.
\end{lstlisting}

After that we translated using our own Neural Machine Translation,

\begin{lstlisting}[breaklines=true]
  Untuk mengubah skrip ini, kita Akkan menggunakan konsep pengulangan 'while' dan memastikan indeks tidak melebihi panjang senarai. Selain itu, kita perlu memastikan skrip kita mengawal senarai kosong dengan betul.

  Berikut adalah cara untuk mengubahnya:

  ```python
    # Set up a list of integers
    arr = [1, 2, 3, 4]

    # Tentukan panjang senarai
    n = len(arr)

    # Pastikan senarai tidak kosong
    if n > 0:
      i = 0
      # Melalui senarai dan keluarkan setiap elemen individu
      while i < n:
          print(arr[i])
          i += 1
    else:
      print("Senarai adalah kosong.")
  ```

  Dalam skrip ini, sebelum kita mula mengulangi, kita memeriksa sama ada `n`, iaitu panjang senarai, lebih besar daripada 0. Jika tidak (yang bermaksud senarai kosong), kita mencetak mesej yang sesuai.

  Kemudian, kita menggunakan pengulangan 'while' untuk mengulangi semasa indeks `i` kurang daripada panjang `n` senarai. Semasa setiap iterasi, kita mencetak elemen senarai dan menambahkan 1 ke indeks, supaya kita boleh bergerak ke elemen seterusnya pada iterasi seterusnya.

  Ini akan mengelakkan ralat indeks, kerana kita tidak akan mencuba mengakses indeks yang melebihi had senarai.
\end{lstlisting}

All synthetic dataset and implementation published at \href{https://huggingface.co/datasets/mesolitica/chatgpt4-code-instruct}{mesolitica/chatgpt4-code-instruct}.

\subsubsection{Synthetic Malay CommonsenseQA}

We expanded our training dataset by creating a set of synthetic common sense questions and answers using ChatGPT4. Taking inspiration from CommonsenseQA \cite{talmor2019commonsenseqa}, we developed questions that explore common sense reasoning. This addition to our training data helps the language model improve its understanding of everyday situations and enhances its ability to generate commonsense responses.
Below is the prompt to generate the synthetic dataset,

\begin{lstlisting}[breaklines=true]
  question = 'The sanctions against the school were a punish...'
  choices = {'label': ['A', 'B', 'C', 'D', 'E']}
  answerKey = 'A'
  instruction = f'Jawab soalan berikut.\n{question}\n{choices}\nJawapan: {answerKey}'
  instruction = f'{instruction}\n\ngenerate similar malay questions included answer'
  chatgpt4(instruction)
\end{lstlisting}

An example of generated dataset,

\begin{lstlisting}[breaklines=true]
  Question: 1. Seseorang yang bersara mungkin perlu kembali bekerja jika mereka apa? A. mempunyai hutang B. mencari pendapatan C. meninggalkan pekerjaan D. memerlukan wang E. kehilangan kunci
  Anwer: D
\end{lstlisting}

All synthetic dataset and implementation published at \href{https://huggingface.co/datasets/mesolitica/chatgpt4-commonsense-qa}{mesolitica/chatgpt4-commonsense-qa}.

\subsubsection{Synthetic Malaysian QA}\label{sec:synthetic-malaysian-qa}

We generated a comprehensive set of questions and answers encompassing 58 diverse Malaysian topics, spanning various domains such as politics, socioeconomy, culture, gender, religion, sociology, social class, technology, ethnicity, infrastructure, health, education, ecology, party politics, diplomacy, history, cuisine, microeconomics, business, artificial intelligence, law, and more. The topics also include specific regions and states such as Negeri Johor, Negeri Kedah, Negeri Kelantan, and others, as well as political parties such as Parti Keadilan Rakyat and Parti Islam SeMalaysia. Additionally, the set covers cultural aspects like literature, language (Tatabahasa, Kesusasteraan Melayu), poetry forms like Pantun, Sajak, and Syair, and socio-political concepts like Constitutional Monarchy, Parliamentary Democracy, and Political Economy. This extensive set of questions and answers aims to provide a nuanced and comprehensive exploration of diverse topics relevant to the Malaysian context.
Below is the prompt to generate the synthetic dataset,

\begin{lstlisting}[breaklines=true]
  topic = 'politics'
  instruction = f'generate random very specific {topic} questions dalam bahasa melayu related to malaysian context'
  chatgpt4(instruction)
\end{lstlisting}

An example of generated dataset,

\begin{lstlisting}[breaklines=true]
{'question': 'Adakah AI boleh digunakan untuk mempercepat proses pengesanan dan rawatan penyakit berjangkit di Malaysia?',
 'answer': 'Ya, AI boleh digunakan untuk mempercepat proses pengesanan dan rawatan penyakit berjangkit di Malaysia. AI dapat membantu dalam pengumpulan dan analisis data kesihatan secara cepat dan tepat, membolehkan doktor dan pakar kesihatan membuat keputusan yang lebih baik dan pantas dalam merawat pesakit. AI juga boleh digunakan untuk mengesan pola penyebaran penyakit berjangkit, memungkinkan tindakan pencegahan dan kawalan yang lebih efektif. Sebagai contoh, AI boleh digunakan dalam sistem pengawasan epidemik untuk mengenal pasti kawasan yang berisiko tinggi dan mengambil langkah-langkah pencegahan segera.'}
\end{lstlisting}

All synthetic dataset and implementation published at \href{https://huggingface.co/datasets/mesolitica/chatgpt4-malaysian-general-qa#synthetic-malaysian-qa}{mesolitica/chatgpt4-malaysian-general-qa}.

\subsubsection{Synthetic Ayat Aktif to Ayat Pasif}

Derived from content available at \url{https://soalanspm.com/ayat-aktif-dan-ayat-pasif/}, we utilized ChatGPT4 to generate synthetic transformations from active sentence (ayat aktif) to passive sentence (ayat pasif). This process involved leveraging the capabilities of ChatGPT4 to produce linguistically accurate and contextually relevant conversions, contributing to the enrichment of our dataset with diverse examples of passive sentence structures derived from the provided active sentences.
Below is the prompt to generate the synthetic dataset,

\begin{lstlisting}[breaklines=true]
  instruction = 'Ayat Aktif: Puan Aishah menasihati Ali agar belajar bersungguh-sungguh kerana peperiksaan hampir tiba.\nAyat Pasif: Ali dinasihati oleh Puan Aishah agar bersungguh-sungguh kerana peperiksaan hampir tiba.'
  instruction = f'{instruction}\n\ngenerate similar questions included answers like above.'
  chatgpt4(instruction)
\end{lstlisting}

An example of generated dataset,

\begin{lstlisting}[breaklines=true]
  {'s': 'Ayat Aktif: Encik Razak mengajar pelajar-pelajar tentang kepentingan menjaga alam sekitar.\nAyat Pasif: Pelajar-pelajar diajar tentang kepentingan menjaga alam sekitar oleh Encik Razak.'}
\end{lstlisting}

All synthetic dataset and implementation published at \href{https://huggingface.co/datasets/mesolitica/chatgpt4-ayat-aktif-pasif}{mesolitica/chatgpt4-ayat-aktif-pasif}.

\subsubsection{Synthetic Malay Paper 1}

We expanded the diversity of our Malay Paper 1 dataset by incorporating synthetic questions generated from various Malay Paper 1 websites. Leveraging the capabilities of ChatGPT4, we prompted the model to generate questions similar to those found in Malay Paper 1 examinations. This approach allowed us to introduce a broader range of question types and structures, enhancing the overall diversity of our dataset. The synthetic questions generated through ChatGPT4 provide additional variations, contributing to a more comprehensive and representative dataset for Malay Paper 1 examinations.
Below is the prompt to generate the synthetic dataset,

\begin{lstlisting}[breaklines=true]
  instruction = 'Jawab soalan berikut.\n3 Manisah amat menggemari pekasam ikan _______ yang dibuat oleh neneknya.\nA peparam\nC pepuyu\nB jejentik\nD reriang\nJawapan: C'
  instruction = f'{instruction}\n\ngenerate similar questions included answers like above.'
\end{lstlisting}

An example of generated dataset,

\begin{lstlisting}[breaklines=true]
  {'question': '1. ........, kamu sudah pandai bermain gitar sekarang!\nA. Oh\nB. Eh\nC. Hai\nD. Ah',
 'answer': 'B'}
\end{lstlisting}

All synthetic dataset and implementation published at \href{https://huggingface.co/datasets/mesolitica/chatgpt4-kertas1}{mesolitica/chatgpt4-kertas1}.

\subsubsection{synthetic Malaysian Open QA Choice}\label{sec:synthetic-qa-choice}

We employed the tool at \url{https://github.com/jxnl/instructor} to generate Open QA Choice questions from diverse sources such as MS Wikipedia, MajalahSains, and Dewan Bahasa articles. Leveraging the Pydantic class, we applied a structured approach to prompt ChatGPT3.5 for generating Open QA Choice responses within the given context. This methodology facilitated the creation of a dataset enriched with Open QA Choice questions derived from authoritative and varied content, offering a more comprehensive and nuanced set of queries for the language model.
Below is the prompt to generate the synthetic dataset,

\begin{lstlisting}[breaklines=true]
  from pydantic import BaseModel, Field
  from enum import Enum
  from typing import List

  class AnswerEnum(str, Enum):
      A = 'A'
      B = 'B'
      C = 'C'
      D = 'D'
      
  class Selective_QA(BaseModel):
      question: str
      A: str
      B: str
      C: str
      D: str
      answer: AnswerEnum
          
  class QAS(BaseModel):
      qa: List[Selective_QA]
  
  paragraph = 'Pelaburan Syarikat China di Malaysia Tingkat Hubungan ...'
  instruction = f"""
  paragraph ```
  {paragraph}
  ```

  berdasarkan paragraph, jana soalan melayu dan jawapan melayu
  """
  chatgpt(instruction, response_model=QAS)
\end{lstlisting}

An example of generated dataset,

\begin{lstlisting}[breaklines=true]
  {'paragraph': 'Pelaburan Syarikat China di Malaysia Tingkat Hubungan ...',
 'qa': {'qa': [{'question': 'Siapakah Menteri Perdagangan Antarabangsa dan Industri Malaysia?',
    'A': 'Tengku Datuk Seri Utama Tengku Zafrul Aziz',
    'B': 'Datuk Arham Abdul Rahman',
    'C': 'Tengku Zafrul Tengku Abdul Aziz',
    'D': 'Datuk Seri Utama Tengku Zafrul Aziz',
    'answer': 'A'},
   {'question': 'Apakah yang dikatakan oleh Tengku Datuk Seri Utama Tengku Zafrul Aziz mengenai peningkatan minat syarikat China melabur di Malaysia?',
    'A': 'Memberi petanda baik kepada negara',
    'B': 'Meningkatkan hubungan dua hala antara Malaysia dan China',
    'C': 'Disokong oleh keyakinan terhadap kerajaan Perpaduan negara',
    'D': 'Semua jawapan di atas betul',
    'answer': 'D'},
   {'question': 'Berapakah nilai pelaburan berpotensi yang diperoleh daripada syarikat China yang bernilai RM 170 bilion?',
    'A': 'RM 55.4 bilion',
    'B': 'RM 264.4 bilion',
    'C': 'RM 170 bilion',
    'D': 'RM 11 545',
    'answer': 'C'},
   {'question': 'Berapakah bilangan projek yang melibatkan China dan telah diluluskan dalam tahun 2022?',
    'A': '91 projek',
    'B': '20 projek',
    'C': '11 545 projek',
    'D': 'Tidak dinyatakan dalam teks',
    'answer': 'A'}]}}
\end{lstlisting}

All synthetic dataset and implementation published at \href{https://huggingface.co/datasets/mesolitica/chatgpt-malaysian-qa-choice}{mesolitica/chatgpt-malaysian-qa-choice}.

\subsubsection{synthetic Malaysian Open QA}

We adopted the approach outlined in ~\ref{sec:synthetic-qa-choice} to generate Open QA questions. This involved extracting content from diverse sources, which are from MS Wikipedia, Malaysia Hansard, and Malay CommonCrawl. By following this approach, we aimed to create a robust dataset of Open QA questions that span a wide spectrum of topics and contexts. The inclusion of content from reputable sources ensures that the questions generated are both informative and relevant, contributing to a comprehensive dataset for training and evaluating language models in the Open QA domain.

Below is the prompt to generate the synthetic dataset,

\begin{lstlisting}[breaklines=true]
  from pydantic import BaseModel, Field
  from typing import List

  class QA(BaseModel):
      question: str
      answer: str
          
  class QAS(BaseModel):
      qa: List[QA]
    
    paragraph = 'PANDAN JAYA SITI AISHAH KUALA LUMPUR SUHANI KUALA LUMPUR ...'
    instruction = f"""
    paragraph ```
    {paragraph}
    ```
  
    berdasarkan paragraph, jana soalan melayu dan jawapan melayu
    """
    chatgpt(instruction, response_model=QAS)
\end{lstlisting}

An example of generated dataset,

\begin{lstlisting}[breaklines=true]
  {'paragraph': "PANDAN JAYA SITI AISHAH KUALA LUMPUR SUHANI KUALA LUMPUR ...",
 'qa': {'qa': [{'question': 'Siapakah yang berada di Pantai Dalam?',
    'answer': 'Suhani'},
   {'question': 'Di mana lokasi SHAH di Alor Gajah?', 'answer': 'Bukit Katil'},
   {'question': 'Siapakah yang berada di Taman Rembau Utama?',
    'answer': 'Mazliena Mohd'}]}}
\end{lstlisting}

All synthetic dataset and implementation published at \href{https://huggingface.co/datasets/mesolitica/chatgpt-malaysian-open-qa}{mesolitica/chatgpt-malaysian-open-qa}.

\subsubsection{Malay Instruction using Alpaca Evolution}

We follow evolution instruction from \cite{xu2023wizardlm} with slightly changes,

Original breadth instruction,

\begin{lstlisting}[]
  I want you act as a Prompt Creator.
  Your goal is to draw inspiration from the #Given Prompt# to create a brand new prompt.
  This new prompt should belong to the same domain as the #Given Prompt# but be even more rare.
  The LENGTH and complexity of the #Created Prompt# should be similar to that of the #Given Prompt#.
  The #Created Prompt# must be reasonable and must be understood and responded by humans.
  '#Given Prompt#', '#Created Prompt#', 'given prompt' and 'created prompt' are not allowed to appear in #Created Prompt#
\end{lstlisting}

Our breadth instruction,

\begin{lstlisting}[]
  I want you act as a Malay Prompt Creator.
  Your goal is to draw inspiration from the #Given Prompt# to create a brand new prompt in malay language and malaysia related if possible.
  This new prompt should belong to the same domain as the #Given Prompt# but be even more rare.
  The LENGTH and complexity of the #Created Prompt# should be similar to that of the #Given Prompt#.
  The #Created Prompt# must be reasonable and must be understood and responded by humans.
  '#Given Prompt#', '#Created Prompt#', 'given prompt' and 'created prompt' are not allowed to appear in #Created Prompt#
\end{lstlisting}

Original depth instruction,

\begin{lstlisting}[]
  I want you act as a Prompt Rewriter.
   Your objective is to rewrite a given prompt into a more complex version to make those famous AI systems (e.g., chatgpt and GPT4) a bit harder to handle.
   But the rewritten prompt must be reasonable and must be understood and responded by humans.
   Your rewriting cannot omit the non-text parts such as the table and code in #The Given Prompt#:. Also, please do not omit the input in #The Given Prompt#. 
   You SHOULD complicate the given prompt using the following method: 
  {} 
  You should try your best not to make the #Rewritten Prompt# become verbose, #Rewritten Prompt# can only add 10 to 20 words into #The Given Prompt#. 
  '#The Given Prompt#', '#Rewritten Prompt#', 'given prompt' and 'rewritten prompt' are not allowed to appear in #Rewritten Prompt#
\end{lstlisting}

Our depth instruction,

\begin{lstlisting}[]
  I want you act as a Malay Prompt Rewriter.
   Your objective is to rewrite a given prompt into malay language, a more complex version and malaysia related if possible to make those famous AI systems (e.g., chatgpt and GPT4) a bit harder to handle.
   But the rewritten prompt must be reasonable and must be understood and responded by humans.
   Your rewriting cannot omit the non-text parts such as the table and code in #The Given Prompt#:. Also, please do not omit the input in #The Given Prompt#. 
   You SHOULD complicate the given prompt using the following method: 
  {} 
  You should try your best not to make the #Rewritten Prompt# become verbose, #Rewritten Prompt# can only add 10 to 20 words into #The Given Prompt#. 
	'#The Given Prompt#', '#Rewritten Prompt#', 'given prompt' and 'rewritten prompt' are not allowed to appear in #Rewritten Prompt#
\end{lstlisting}

Our approach involved utilizing the dataset available at \href{https://raw.githubusercontent.com/gururise/AlpacaDataCleaned/main/alpaca_data_cleaned.json}{AlpacaDataCleaned} as a starting point. To refine and enhance this dataset, we applied the Evolution Instruction above using ChatGPT3.5. Once we obtained the synthetic instructions, we employed ChatGPT3.5 to generate corresponding answers. In cases where the answers were generated in Indonesian, we employed the Neural Machine Translation mentioned at ~\ref{sec:generating-instruction-dataset} to translate them into Malay. This multi-step process aimed to produce a dataset with improved diversity and accuracy, ensuring that the generated questions and answers are not only contextually relevant but also linguistically aligned with the Malaysian language.
An example of generated dataset,

\begin{lstlisting}[breaklines=true]
  Instruction: Tulislah sebuah cerita pendek dengan gaya narasi orang ketiga tentang seorang protagonis yang harus membuat keputusan karier penting. Ceritakan bagaimana protagonis tersebut menghadapi dilema antara mempertahankan pekerjaan yang stabil atau mengambil risiko dengan mencari peluang karier baru di Malaysia yang menjanjikan kehidupan yang lebih baik. Kembangkan karakter dan latar belakang protagonis, serta ceritakan perjalanan dan pertimbangannya dalam mengambil keputusan penting tersebut.
  Answer: Di sebuah bandar kecil di Indonesia, hiduplah seorang wanita bernama Maya. Maya adalah seorang pekerja keras yang telah bekerja di sebuah syarikat besar selama lima tahun. Dia mempunyai pekerjaan yang stabil dan gaji yang mencukupi. Namun, dalam hatinya, Maya merasakan bahawa pekerjaannya tidak memberikan kepuasan dan cabaran yang mencukupi untuknya. Maya bercita-cita untuk meningkatkan kerjaya dan mencapai kejayaan yang lebih besar. Dia sering bermimpi untuk hidup yang lebih baik, dengan pendapatan yang lebih tinggi dan peluang untuk mengembara ke seluruh dunia. Pada suatu hari, Maya mendengar berita bahawa di Malaysia terdapat peluang kerjaya yang menjanjikan. Maya menjadi tertarik dengan peluang ini. Dia mula mencipta tentang industri dan peluang pekerjaan di Malaysia. Maya tahu bahawa kerjaya di sana boleh memberikan pendapatan yang jauh lebih tinggi daripada yang dia terima sekarang. Selain itu, dia juga tahu bahawa di Malaysia terdapat persekitaran kerja yang lebih dinamik dan berbeza-beza, yang akan memberikan cabaran baru untuknya. Namun, Maya juga menyedari bahawa untuk mencari peluang kerjaya di Malaysia, dia harus meninggalkan semua yang dia telah bangunkan di Indonesia, termasuk keluarga dan rakan-rakannya. Maya berasa sedih dan terfikir, kerana dia sangat menyayangi keluarga dan tidak tahan berpisah dengan mereka. Maya menghabiskan masa berhari-hari memikirkan pilihan kerjaya ini. Dia menulis segala kelebihan dan kekurangan kedua-dua pilihan yang dia hadapi. Dia bercakap dengan keluarga dan rakan-rakannya, dan mereka menyokong keputusannya, walaupun mereka sedih dengan pemergian Maya. Akhirnya, selepas pertimbangan yang teliti, Maya memutuskan untuk mengambil risiko dan mencari peluang kerjaya baru di Malaysia. Dia merasakan bahawa ini adalah peluang yang jarang berlaku, dan dia harus mengambil langkah di luar zon selesa untuk mencapai impian dan cita-citanya. Maya menyedari bahawa hidup adalah tentang mengambil risiko dan menghadapi cabaran baru. Perjalanan Maya ke Malaysia bukanlah mudah. Dia harus melalui beberapa halangan dan penyesuaian budaya baru. Namun, dengan ketabahan dan semangat yang kuat, Maya berjaya mendapatkan pekerjaan di sebuah syarikat besar di Kuala Lumpur. Dia berasa gembira dan bangga dengan keputusannya. Selama bertahun-tahun di Malaysia, Maya mencapai kejayaan yang luar biasa. Dia menikmati kerjaya yang dia lalui dan mempunyai kehidupan yang lebih baik dan lebih selesa daripada yang pernah dia bayangkan di Indonesia. Maya juga terus menjalin hubungan dengan keluarga dan rakan-rakannya di negara ini, walaupun jarak memisahkan mereka. Dalam cerita ini, Maya adalah contoh yang memberi inspirasi kepada seseorang yang berani mengambil risiko dengan meninggalkan pekerjaan yang stabil dan mencari peluang kerjaya baru. Keputusannya membawa perubahan besar dalam hidupnya, dan ia menunjukkan kepada kita bahawa kadang-kadang kita harus berani keluar dari zon selesa untuk mencapai impian dan cita-cita kita.
\end{lstlisting}

Additionally, we replicated these steps on ~\ref{sec:synthetic-malaysian-qa}, introducing additional perspectives and topics to create a more comprehensive and inclusive collection of synthetic questions and answers from the base topics.
An example of generated dataset,

\begin{lstlisting}[breaklines=true]
  {'instruction': 'Dapatkah anda membuat suatu versi yang diperbaharui daripada ayat yang memberi fokus kepada kepentingan kecocokan budaya dalam penilaian calon? \n\nMemang, saya memahami tanggungjawab jawatan ini dan bersedia untuk menganalisis serta mengkaji kecekapan teknikal dan kebolehan komunikasi calon dengan cekap, memberikan maklum balas yang bernas terhadap jawapan mereka, dan memberikan cadangan yang mendalam berkenaan kesesuaian mereka dengan jawatan tersebut.',
 'output': 'Tentu, berikut ini adalah versi yang diperbaharui daripada ayat yang lebih memberi fokus kepada kepentingan kecocokan budaya dalam penilaian calon:\n\n"Saya menaruh perhatian yang besar terhadap pentingnya kecocokan budaya dalam menilai calon. Saya sepenuhnya memahami tanggung jawab jabatan ini dan siap untuk menganalisis dan mengevaluasi secara teliti keahlian teknis dan kemampuan komunikasi calon. Saya akan memberikan tanggapan yang mendalam terhadap jawaban mereka dan memberikan saran yang relevan mengenai kesesuaian mereka dengan jabatan ini."',
 'indon_ins': False,
 'indon_output': True,
 'instruction_ms': 'Bolehkah anda membuat versi yang diperbaharui daripada ayat yang memberi tumpuan kepada kepentingan kecocokan budaya dalam penilaian calon?\n\nMemang, saya memahami tanggungjawab jawatan ini dan bersedia untuk menganalisis serta mengkaji kecekapan teknikal dan kebolehan komunikasi calon dengan cekap, memberikan maklum balas yang bernas terhadap jawapan mereka, dan memberikan cadangan yang mendalam berkenaan kesesuaian mereka dengan jawatan tersebut.',
 'output_ms': 'Sudah tentu, berikut adalah versi yang diperbaharui daripada ayat yang lebih memberi tumpuan kepada kepentingan kesesuaian budaya dalam penilaian calon:\n\n"Saya memberi perhatian yang besar kepada kepentingan kesesuaian budaya dalam menilai calon. Saya sepenuhnya memahami tanggungjawab jawatan ini dan bersedia untuk menganalisis dan menilai dengan teliti kepakaran teknikal dan keupayaan komunikasi calon. Saya akan memberikan jawapan yang mendalam terhadap jawapan mereka dan memberikan nasihat yang relevan tentang kesesuaian mereka dengan jawatan ini."',
 'rejected_ins': False,
 'rejected_output': False}
\end{lstlisting}

All synthetic dataset and implementation published at \href{https://huggingface.co/datasets/mesolitica/chatgpt-malay-instructions}{mesolitica/chatgpt-malay-instructions}.

\subsubsection{Malay UltraChat}\label{sec:malay-ultrachat}

Our objective extends beyond single-turn datasets, as we aim to generate a diverse multiturn dataset encompassing various sources such as Astro Awani, papers with the 'melayu' keyword from Crossref, Epenerbitan, public PDFs from gov.my, JurnalDBP, lom.agc.gov.my, MS Wikipedia, Malaysia Hansard, textbooks, \url{https://maktabahalbakri.com/}, and \url{https://muftiwp.gov.my/ms/}. Inspired by the approach taken by Ultrachat \cite{ding2023enhancing}, we seek to create a robust and multifaceted dataset that captures the intricacies of conversational exchanges across a broad spectrum of topics, sources, and contexts. This endeavor contributes to the development of language models capable of handling multiturn conversations with depth and relevance, reflecting real-world conversational dynamics.

The first step to generate UltraChat, we must generate list of questions, below is the prompt we use to generate the questions,

\begin{lstlisting}[breaklines=true]
  paragraph = '17 Oktober 2014 \n17 October 2014\n\nP.U. (A) 279   WARTA KERAJAAN PERSEKUTUAN\n\n ...'
  instruction = f'{paragraph}\ngenerate a question related to the context'
  chatgpt(instruction)
\end{lstlisting}

In the next step, we simply loop through the dataset N times, continually adding the generated answers to the initial context,

\begin{lstlisting}[breaklines=true]
  system_malay = 'Above is a conversation between a user and an intelligent assistant. You suppose to simulate conversation between the user and the assistant in malay language. Bear in mind your major request is to ask the assistant to generate some material. So you can ask the assistant either to make it more detailed, add more related information, or any other request to improve the generated material. Be creative and diverse in your request. Make the response short and the language casual.'
  def ultrachat(row, n = 1):
    results = [
        {'role': 'context', 'content': row['paragraph']},
        {'role': 'user', 'content': row['question']},
    ]
    initial = f"""
    {row['paragraph']}

    {row['question']}
    """.strip()
    messages = [
        {'role': 'system', 'content': system_malay},
        {'role': 'user', 'content': initial},
    ]
    r = chatgpt(messages)
    results.append({
        'role': 'assistant', 'content': r,
    })
    messages.append({
        'role': 'assistant', 'content': r,
    })
    
    for _ in range(n):
        messages_temp = messages + [
            {'role': 'user', 'content': 'Now suppose you are the user, say something to continue the conversation based on given context. Make the response short and the language casual'}
        ]
        r = chatgpt(messages_temp)
        results.append({
            'role': 'user', 'content': r,
        })
        messages.append({
            'role': 'user', 'content': r,
        })
        r = chatgpt(messages)
        results.append({
            'role': 'assistant', 'content': r,
        })
        messages.append({
            'role': 'assistant', 'content': r,
        })
    
    return results
\end{lstlisting}

An example of generated dataset,

\begin{lstlisting}[breaklines=true]
  [{'role': 'context',
  'content': 'LIGA\xa0Malaysia (Liga-M) terus menerima tamparan hebat apabila UiTM FC sah menarik diri daripada saingan Liga Super musim ini bagi menyusuli jejak Petaling Jaya City.\n\n\n\r\nPerkara itu disahkan sendiri oleh Presiden The Lion Troops, Profesor Dr. Sazili Shahibi, selepas dua musim beraksi di pentas Liga Super.\n\n\n\r\nSelain masalah kewangan, Dr Sazili menjelaskan ada punca lain mengapa pihak pengurusan mengibarkan bendera putih.\n\n\n \n \n\r\n"Ahli Lembaga Pengarah UiTM telah bermesyuarat dan (kami mengambil) keputusan untuk menarik diri daripada liga pada musim depan.\n\n\n\r\n"Kewangan adalah salah faktornya, tapi kami rasa daripada segi kemampuan, ia tidak sesuai buat pemain-pemain kami," katanya ketika dihubungi \nAstro Arena\n\xa0pada Rabu.\n\n\n\r\nSazili juga berhasrat memaklumkan UiTM bakal beralih ke saingan Liga M3 musim ini.\n\n\n\r\nUntuk rekod, UiTM muncul pasukan berstatus universiti pertama yang layak beraksi dalam Liga Super.\n\n\n\r\nDalam pada itu, Liga Bolasepak Malaysia (MFLhanya akan menjelaskan situasi berkenaan menerusi kenyataan rasmi bakal dalam masa terdekat selepas hanya 14 pasukan yang akan bertanding di pentas Liga Super musim ini. -\nStadium Astro',
  'content_ms': None,
  'indon': False},
 {'role': 'user',
  'content': 'What are the reasons given by UiTM FC for withdrawing from Liga Malaysia?',
  'content_ms': 'Apakah alasan yang diberikan oleh UiTM FC kerana menarik diri daripada Liga Malaysia?',
  'indon': False},
 {'role': 'assistant',
  'content': 'Menurut Presiden UiTM FC, Profesor Dr. Sazili Shahibi, mereka mengambil keputusan untuk menarik diri daripada Liga Malaysia musim ini atas beberapa sebab. Salah satu sebab utamanya adalah masalah kewangan yang dihadapi oleh pasukan. Selain itu, beliau juga menyebut bahawa kemampuan pasukan UiTM tidak sesuai untuk bersaing dalam Liga Super. Selain masalah ini, tidak ada sebab lain yang diberikan dalam laporan tersebut.',
  'content_ms': 'Menurut Presiden UiTM FC, Profesor Dr. Sazili Shahibi, mereka membuat keputusan untuk menarik diri daripada Liga Malaysia musim ini atas beberapa sebab. Salah satu sebab utama ialah masalah kewangan yang dihadapi oleh pasukan. Selain itu, beliau juga menyatakan bahawa keupayaan pasukan UiTM tidak sesuai untuk bersaing dalam Liga Super. Selain masalah ini, tiada sebab lain diberikan dalam laporan tersebut.',
  'indon': False},
 {'role': 'user',
  'content': 'Alamak, memang sedihlah tengok UiTM FC tarik diri dari Liga Malaysia. Macam mana pula dengan penonton setia yang selalu sokong pasukan tu? Harap-harap ada jalan penyelesaian untuk masalah kewangan mereka.',
  'content_ms': 'Alamak, memang sedih tengok UiTM FC menarik diri dari Liga Malaysia. Bagaimana pula dengan penonton setia yang selalu menyokong pasukan itu? Harap-harap ada jalan penyelesaian untuk masalah kewangan mereka.',
  'indon': False},
 {'role': 'assistant',
  'content': 'Ya, memang sedih melihat UiTM FC menarik diri dari Liga Malaysia. Bagi penonton setia yang sentiasa menyokong pasukan tersebut, tentu mereka juga berasa kecewa. Kami juga berharap ada jalan penyelesaian untuk masalah kewangan yang dihadapi oleh UiTM FC. Mungkin pihak pengurusan boleh mencari sponsor atau menyusun semula strategi kewangan untuk membolehkan pasukan tersebut terus beraksi dalam liga pada masa hadapan. Semoga mereka dapat menemui penyelesaian yang baik dan kembali bersaing di pentas bola sepak Malaysia.',
  'content_ms': 'Ya, memang sedih melihat UiTM FC menarik diri dari Liga Malaysia. Bagi penonton setia yang sentiasa menyokong pasukan tersebut, tentu mereka juga berasa kecewa. Kami juga berharap ada jalan penyelesaian untuk masalah kewangan yang dihadapi oleh UiTM FC. Mungkin pihak pengurusan boleh mencari penaja atau menyusun semula strategi kewangan untuk membolehkan pasukan tersebut terus beraksi dalam liga pada masa hadapan. Semoga mereka dapat menemui penyelesaian yang baik dan kembali bersaing di pentas bola sepak Malaysia.',
  'indon': False},
 {'role': 'user',
  'content': 'Ya betul tu, banyak peminat setia UiTM FC yang akan kecewa dengan keputusan mereka tarik diri dari Liga Malaysia. Harapnya pihak pengurusan dapat cari jalan penyelesaian secepat mungkin supaya pasukan tu masih dapat beraksi dalam liga. Saya pun tak sabar nak tengok prestasi UiTM FC lagi musim hadapan. Semoga ada berita baik dalam masa terdekat.',
  'content_ms': 'Ya, betul tu. Banyak peminat setia UiTM FC yang akan kecewa dengan keputusan mereka menarik diri dari Liga Malaysia. Harapnya pihak pengurusan dapat mencari jalan penyelesaian secepat mungkin supaya pasukan itu masih dapat beraksi dalam liga. Saya pun tak sabar nak tengok prestasi UiTM FC lagi musim depan. Semoga ada berita baik dalam masa terdekat.',
  'indon': False},
 {'role': 'assistant',
  'content': 'Saya juga setuju dengan anda. Semoga pihak pengurusan UiTM FC dapat mencari jalan penyelesaian secepat mungkin agar pasukan tersebut masih dapat berlaga dalam Liga Malaysia. Benar, banyak peminat yang ingin melihat prestasi UiTM FC musim hadapan. Mari kita nantikan berita baik daripada mereka dalam masa terdekat. Harapnya mereka dapat mengatasi masalah kewangan dan kembali membanggakan para penyokong. Kita terus sokong UiTM FC!',
  'content_ms': 'Saya juga bersetuju dengan anda. Semoga pihak pengurusan UiTM FC dapat mencari jalan penyelesaian secepat mungkin supaya pasukan tersebut masih dapat bersaing dalam Liga Malaysia. Benar, ramai peminat yang ingin melihat prestasi UiTM FC musim depan. Mari kita nantikan berita baik daripada mereka dalam masa terdekat. Harapnya mereka dapat mengatasi masalah kewangan dan kembali membanggakan para penyokong. Kita terus sokong UiTM FC!',
  'indon': False},
 {'role': 'user',
  'content': 'Ya lah, harapnya UiTM FC dapat dapat mencari penyelesaian yang cepat. Penyokong setia UiTM FC mesti excited nak tengok pasukan beraksi lagi musim hadapan. Semoga ada berita gembira dalam masa terdekat!',
  'content_ms': 'Ya lah, harapnya UiTM FC dapat mencari penyelesaian yang cepat. Penyokong setia UiTM FC mesti teruja untuk melihat pasukan beraksi lagi musim depan. Semoga ada berita gembira dalam masa terdekat!',
  'indon': False},
 {'role': 'assistant',
  'content': 'Betul tu, penyokong setia UiTM FC pasti sangat teruja melihat pasukan beraksi di musim hadapan. Kita harap pihak pengurusan dapat mencari penyelesaian dengan pantas supaya pasukan dapat kembali ke gelanggang. Kita tunggu berita gembira yang akan datang dan terus menyokong UiTM FC dengan penuh semangat. Biar bola sepak Malaysia sentiasa berwarna dengan kehadiran pasukan hebat seperti UiTM FC!',
  'content_ms': 'Betul tu, penyokong setia UiTM FC pasti sangat teruja melihat pasukan beraksi pada musim depan. Kita harap pihak pengurusan dapat mencari penyelesaian dengan pantas supaya pasukan dapat kembali ke gelanggang. Kita tunggu berita gembira yang akan datang dan terus menyokong UiTM FC dengan penuh semangat. Biar bola sepak Malaysia sentiasa berwarna dengan kehadiran pasukan hebat seperti UiTM FC!',
  'indon': False},
 {'role': 'user',
  'content': 'Ya betul tu, memang berharap pasukan UiTM FC dapat segera mencari jalan penyelesaian. Setiap pasukan memang perlu sokongan dan kami sebagai penyokong setia akan terus memberikan sokongan kepada pasukan kesayangan kami. Semoga ada berita gembira yang akan datang!',
  'content_ms': 'Ya, betul tu. Memang berharap pasukan UiTM FC dapat segera mencari jalan penyelesaian. Setiap pasukan memang perlukan sokongan dan kami sebagai penyokong setia akan terus memberikan sokongan kepada pasukan kesayangan kami. Semoga ada berita gembira yang akan datang!',
  'indon': False},
 {'role': 'assistant',
  'content': 'Setiap pasukan memang memerlukan sokongan daripada penyokong setia seperti anda. Anda adalah aset berharga bagi pasukan UiTM FC. Teruskan menyokong dengan penuh semangat dan harapan. Kita berdoa agar pihak pengurusan dapat mencari jalan penyelesaian dengan segera. Bersama-sama, mari kita nantikan berita gembira yang akan datang dan terus memberikan sokongan kepada pasukan kesayangan kita. UiTM FC pasti akan bangkit dan menyampaikan prestasi yang cemerlang musim hadapan!',
  'content_ms': 'Setiap pasukan memang memerlukan sokongan daripada penyokong setia seperti anda. Anda adalah aset berharga bagi pasukan UiTM FC. Teruskan menyokong dengan penuh semangat dan harapan. Kita berdoa agar pihak pengurusan dapat mencari jalan penyelesaian dengan segera. Bersama-sama, mari kita nantikan berita gembira yang akan datang dan terus memberikan sokongan kepada pasukan kesayangan kita. UiTM FC pasti akan bangkit dan menyampaikan prestasi yang cemerlang musim depan!',
  'indon': False}]
\end{lstlisting}

To address instances where the output from ChatGPT3.5 is occasionally generated in Indonesian, we use the Neural Machine Translation mentioned at ~\ref{sec:generating-instruction-dataset} to ensure that the answers are consistently in Malay. This additional step helps maintain linguistic coherence and alignment with our dataset's intended language, enhancing the overall quality and consistency of the generated responses.

All synthetic dataset and implementation published at \href{https://huggingface.co/datasets/mesolitica/malaysian-ultrachat}{mesolitica/malaysian-ultrachat}.

\subsubsection{Malay Coding UltraChat}

Derived from the source at \url{https://huggingface.co/datasets/TokenBender/glaive_coder_raw_text}, we employed the same method from ~\ref{sec:malay-ultrachat} to systematically generate a multiturn dataset focused on Malay coding. This approach allowed us to adapt existing content and structure it into a format suitable for multiturn conversations related to coding scenarios. By utilizing the Ultrachat template, we ensured consistency in the style and context of the generated coding interactions, contributing to the development of a specialized multiturn dataset tailored to the domain of Malay coding.

An example of generated dataset,

\begin{lstlisting}[breaklines=true]
  [{'role': 'user',
  'content': 'Bagaimana cara saya menghasilkan teks tebal dalam Bash? Saya mempunyai skrip Bash yang mencetak beberapa teks ke skrin menggunakan perintah `echo "Beberapa Teks"`. Adakah cara untuk memformat teks tersebut supaya ia menjadi tebal?',
  'content_ms': None,
  'indon': False},
 {'role': 'assistant',
  'content': 'Anda boleh menggunakan perintah `echo -e` untuk memformat teks tersebut supaya menjadi tebal. Contohnya, `echo -e "\\033[1mTeks Tebal\\033[0m"`. Dalam contoh ini, `\\033[1m` akan membuat teks menjadi tebal, manakala `\\033[0m` digunakan untuk menetapkan semula pemformatan teks. Semoga membantu!',
  'content_ms': 'Anda boleh menggunakan perintah `echo -e` untuk memformat teks supaya ia menjadi tebal. Contohnya, `echo -e "033[1mTeks Tebal033[0m"`. Dalam contoh ini, `033[1m` akan menjadikan teks tebal, manakala `033[0m` digunakan untuk menetapkan semula format teks. Semoga ia membantu!',
  'indon': False},
 {'role': 'user',
  'content': 'Bagaimana dengan pemformatan teks yang lain? Adakah terdapat cara untuk menjadikannya miring atau bergaris bawah dalam skrip Bash?',
  'content_ms': 'Bagaimana pula dengan format teks lain? Adakah terdapat cara untuk menjadikannya condong atau bergaris bawah dalam skrip Bash?',
  'indon': False},
 {'role': 'assistant',
  'content': 'Tentu! Anda juga boleh memformat teks agar menjadi miring atau bergaris bawah dalam skrip Bash. Untuk menjadikannya miring, anda boleh menggunakan perintah `echo -e "\\033[3mTeks Miring\\033[0m"`. Manakala, untuk menjadikannya bergaris bawah, gunakan perintah `echo -e "\\033[4mTeks Bergaris Bawah\\033[0m"`. Sekali lagi, `\\033[0m` digunakan untuk menetapkan semula pemformatan teks. Semoga ini membantu anda!',
  'content_ms': 'Sudah tentu! Anda juga boleh memformat teks untuk menjadi miring atau bergaris bawah dalam skrip Bash. Untuk menjadikannya miring, anda boleh menggunakan perintah `echo -e "033[3mTeks Miring033[0m"`. Manakala, untuk menjadikannya bergaris bawah, gunakan perintah `echo -e "033[4mTeks Bergaris Bawah033[0m"`. Sekali lagi, `033[0m` digunakan untuk menetapkan semula format teks. Semoga ini membantu anda!',
  'indon': False},
 {'role': 'user',
  'content': 'Wah, sangat menarik! Adakah terdapat juga cara untuk menyertakan warna dalam pemformatan teks dalam skrip Bash?',
  'content_ms': 'Wah, sangat menarik! Adakah terdapat juga cara untuk memasukkan warna dalam format teks dalam skrip Bash?',
  'indon': False},
 {'role': 'assistant',
  'content': 'Ya, anda juga boleh menyertakan warna dalam pemformatan teks dalam skrip Bash. Terdapat beberapa kod warna yang boleh anda gunakan. Sebagai contoh, untuk meletakkan teks dalam warna merah, anda boleh menggunakan perintah `echo -e "\\033[31mTeks Merah\\033[0m"`. Di sini, `\\033[31m` akan mengatur warna menjadi merah dan `\\033[0m` digunakan untuk menetapkan semula pemformatan teks.\n\nBerikut adalah beberapa kod warna yang boleh digunakan:\n- Merah: `\\033[31m`\n- Hijau: `\\033[32m`\n- Kuning: `\\033[33m`\n- Biru: `\\033[34m`\n- Ungu: `\\033[35m`\n- Cyan: `\\033[36m`\n\nAnda juga boleh menggabungkan pemformatan, sebagai contoh, `echo -e "\\033[1;31mTeks Merah Tebal\\033[0m"` akan menghasilkan teks berwarna merah yang tebal. Selamat mencuba!',
  'content_ms': 'Ya, anda juga boleh menyertakan warna dalam format teks dalam skrip Bash. Terdapat beberapa kod warna yang boleh anda gunakan. Sebagai contoh, untuk meletakkan teks dalam warna merah, anda boleh menggunakan perintah `echo -e "033[31mTeks Merah033[0m"`. Di sini, `033[31m` akan menetapkan warna kepada merah dan `033[0m` digunakan untuk menetapkan semula format teks.\n\nBerikut adalah beberapa kod warna yang boleh digunakan:\n- Merah: `033[31m`\n- Hijau: `033[32m`\n- Kuning: `033[33m`\n- Biru: `033[34m`\n- Ungu: `033[35m`\n- Cyan: `033[36m`\n\nAnda juga boleh menggabungkan format, sebagai contoh, `echo -e "033[1;31mTeks Merah Tebal033[0m"` akan menghasilkan teks merah tebal. Selamat mencuba!',
  'indon': False},
 {'role': 'user',
  'content': 'Wah, sungguh menarik! Adakah terdapat kod warna lain yang boleh digunakan dalam pemformatan teks dalam skrip Bash? Saya ingin belajar lebih banyak tentang pemformatan teks ini.',
  'content_ms': 'Wah, sungguh menarik! Adakah terdapat kod warna lain yang boleh digunakan dalam format teks dalam skrip Bash? Saya ingin mengetahui lebih lanjut tentang format teks ini.',
  'indon': False},
 {'role': 'assistant',
  'content': 'Ya, terdapat banyak lagi kod warna yang boleh digunakan dalam pemformatan teks dalam skrip Bash. Berikut adalah beberapa kod warna tambahan untuk variasi yang lebih menarik:\n\n- Putih: `\\033[0m` atau `\\033[97m`\n- Abu-abu Terang: `\\033[37m`\n- Abu-abu Gelap: `\\033[90m`\n- Kuning Terang: `\\033[93m`\n- Merah Terang: `\\033[91m`\n- Biru Terang: `\\033[94m`\n- Hijau Terang: `\\033[92m`\n- Ungu Terang: `\\033[95m`\n- Cyan Terang: `\\033[96m`\n\nSelain itu, anda juga boleh menggabungkan beberapa pemformatan sekaligus untuk mencipta lebih banyak variasi. Misalnya, `echo -e "\\033[1;31;42mTeks Tebal Merah dengan Latar Belakang Hijau\\033[0m"` akan menghasilkan teks yang tebal, berwarna merah, dan dengan latar belakang hijau.\n\nJangan ragu untuk mencoba dan bereksperimen dengan pemformatan teks ini. Selamat mencuba, dan semoga anda menemui hasil yang menarik!',
  'content_ms': 'Ya, terdapat banyak lagi kod warna yang boleh digunakan dalam format teks dalam skrip Bash. Berikut adalah beberapa kod warna tambahan untuk variasi yang lebih menarik:\n\n- Putih: `033[0m` atau `033[97m`\n- Abu-abu Terang: `033[37m`\n- Abu-abu Gelap: `033[90m`\n- Kuning Terang: `033[93m`\n- Merah Terang: `033[91m`\n- Biru Terang: `033[94m`\n- Hijau Terang: `033[92m`\n- Ungu Terang: `033[95m`\n- Cyan Terang: `033[96m`\n\nSelain itu, anda juga boleh menggabungkan beberapa format sekaligus untuk mencipta lebih banyak variasi. Contohnya, `echo -e "033[1;31;42mTeks Tebal Merah dengan Latar Belakang Hijau033[0m"` akan menghasilkan teks yang tebal, merah, dan dengan latar belakang hijau.\n\nJangan ragu untuk mencuba dan bereksperimen dengan format teks ini. Selamat mencuba, dan semoga anda menemui hasil yang menarik!',
  'indon': True},
 {'role': 'user',
  'content': 'Sungguh menarik! Terdapat begitu banyak variasi dan pilihan dalam pemformatan teks di Bash. Saya ingin tahu apakah perbedaan antara "\\033[0m" dan "\\033[39m" dalam pemformatan warna teks? Bisakah Anda jelaskan?',
  'content_ms': 'Sungguh menarik! Terdapat banyak variasi dan pilihan dalam format teks dalam Bash. Saya ingin tahu apakah perbezaan antara "033[0m" dan "033[39m" dalam format warna teks? Bolehkah anda menerangkannya?',
  'indon': False},
 {'role': 'assistant',
  'content': 'Tentu! Perbedaan antara "\\033[0m" dan "\\033[39m" dalam pemformatan warna teks adalah sebagai berikut:\n\n- "\\033[0m" digunakan untuk menetapkan semula pemformatan teks ke default, termasuk warna teks.\n- "\\033[39m" khusus digunakan untuk menetapkan warna teks kembali ke warna default yang biasanya digunakan untuk teks normal.\n\nDalam praktiknya, kedua kode ini biasanya menghasilkan hasil yang sama, yakni mengembalikan warna teks ke default. Namun, ada beberapa sistem atau tampilan terminal yang mungkin memberikan perbedaan kecil di antara keduanya.\n\nJadi, jika anda ingin menetapkan warna teks kembali ke warna default dalam pemformatan teks Bash, anda boleh menggunakan salah satu dari kedua kode tersebut. Semoga ini membantu menjelaskan perbedaan di antara keduanya!',
  'content_ms': 'Sudah tentu! Perbezaan antara "033[0m" dan "033[39m" dalam format warna teks adalah seperti berikut:\n\n- "033[0m" digunakan untuk menetapkan semula format teks kepada lalai, termasuk warna teks.\n- "033[39m" khusus digunakan untuk menetapkan warna teks kembali kepada warna lalai yang biasanya digunakan untuk teks biasa.\n\nDalam praktiknya, kedua-dua kod ini biasanya menghasilkan hasil yang sama, iaitu mengembalikan warna teks kepada lalai. Walau bagaimanapun, terdapat beberapa sistem atau paparan terminal yang mungkin memberikan perbezaan kecil antara kedua-duanya.\n\nJadi, jika anda ingin menetapkan warna teks kembali kepada warna lalai dalam format teks Bash, anda boleh menggunakan salah satu daripada kedua-dua kod. Semoga ini membantu menjelaskan perbezaan antara kedua-duanya!',
  'indon': True}]
\end{lstlisting}

All synthetic dataset and implementation published at \href{https://huggingface.co/datasets/mesolitica/chatgpt-code-instruct-multiturn}{mesolitica/chatgpt-code-instruct-multiturn}.

\pagebreak

\subsection{Malay Instruction Dataset}

\subsubsection{Synthetic Kesalahan Tatabahasa (Grammar Errors) Dataset}

We crafted a synthetic dataset focusing on introducing grammatical errors using the \href{https://malaya.readthedocs.io/en/stable/load-dependency.html}{Malaya dependency parser}. We performed a positional swap based on the parsed tree structure. This deliberate introduction of errors was intended to create a dataset challenging language models to identify and rectify grammatical mistakes.

\begin{figure}[h]
  \centering
  \includegraphics[width=0.6\linewidth]{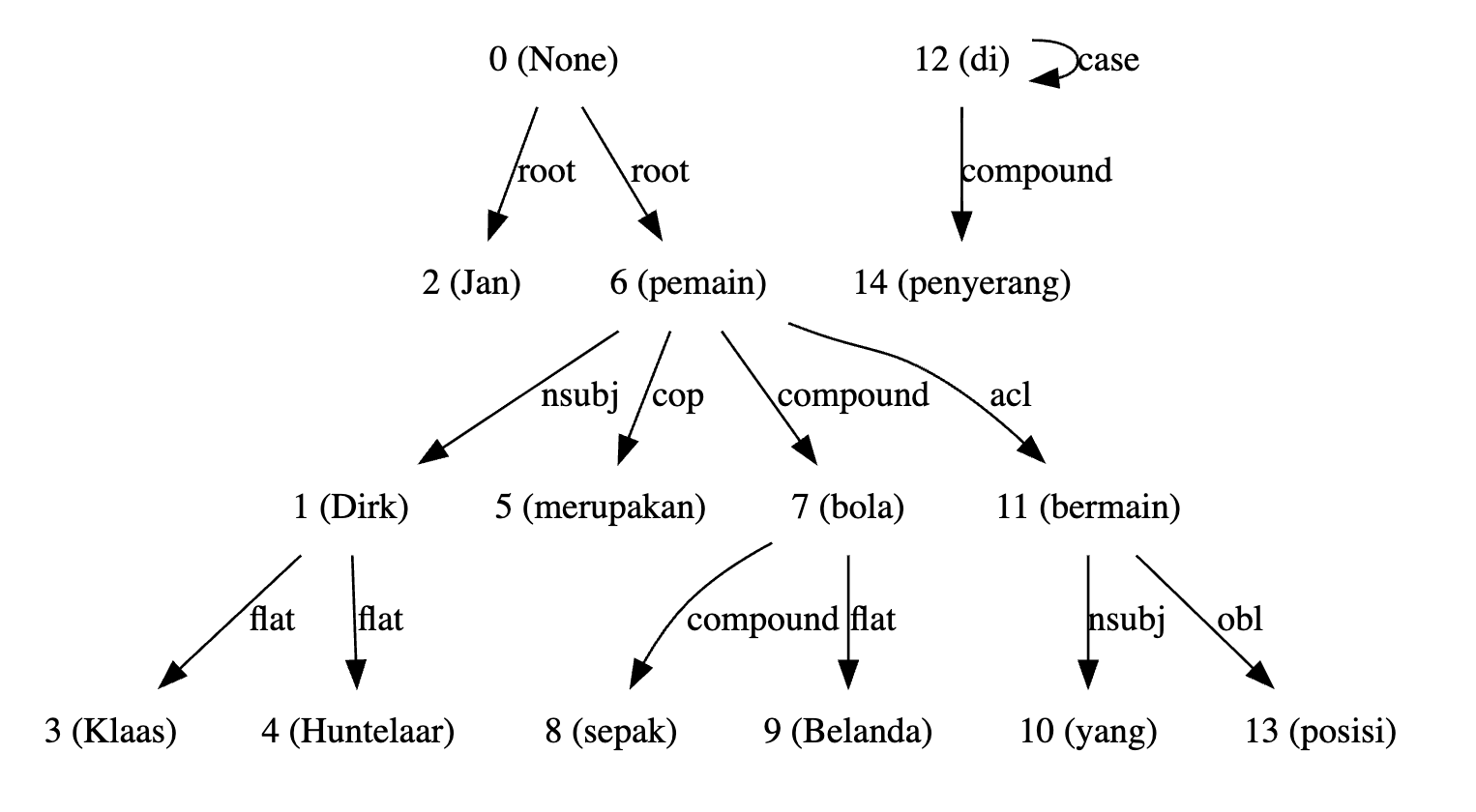} % Replace with your image file
\end{figure}

Based on the figure above, we can exchange the positions of 7 and 8 to introduce compound grammar error. In this \href{https://github.com/mesolitica/malaya/tree/master/session/tatabahasa/prepare-dataset}{malaya/session/tatabahasa} repository, we generated up to 14 different synthetic grammatical errors.

Subsequently, we transformed this dataset into a question-answer format, contributing to training models on error detection and correction in natural language,

\begin{lstlisting}[breaklines=true]
  Context: Ia dirobohkan (bagi) 2005 dan digantikan kepada Hypo-Arena yang segar . agaknya dikenali sehingga 30 Jun 2007 dengan sebutan " Wortherseestadion " :
  Question: Apakah kesalahan tatabahasa untuk (bagi)
  Choice: { "A": "kesalahan kata kerja transitif", "B": "kesalahan kata tanya", "C": "kesalahan kata sendi", "D": "kesalahan kata ganti diri" }
  Answer: C
  Fix: pada
\end{lstlisting}

All synthetic dataset and implementation published at \href{https://huggingface.co/datasets/mesolitica/kesalahan-tatabahasa-choice}{mesolitica/kesalahan-tatabahasa-choice}.

\subsection{Fine-tuning Phase}

During the fine-tuning phase, we implemented full parameter tuning using the Mistral chat template. This process involved an extensive adjustment of all model parameters to optimize the chat interaction performance for malaysian contexts. Figure below shows the example of formatted chat template:

\begin{lstlisting}[breaklines=true]
"<s>[INST] Bagaimana cara saya menghasilkan teks tebal dalam Bash? Saya mempunyai skrip Bash yang mencetak beberapa teks ke skrin menggunakan perintah `echo "Beberapa Teks"`. Adakah cara untuk memformat teks tersebut supaya ia menjadi tebal? [/INST] Anda boleh menggunakan perintah `echo -e` untuk memformat teks supaya ia menjadi tebal. Contohnya, `echo -e "033[1mTeks Tebal033[0m"`. Dalam contoh ini, `033[1m` akan menjadikan teks tebal, manakala `033[0m` digunakan untuk menetapkan semula format teks. Semoga ia membantu!</s> [INST] Bagaimana pula dengan format teks lain? Adakah terdapat cara untuk menjadikannya condong atau bergaris bawah dalam skrip Bash? [/INST]"
\end{lstlisting}

We fine-tuned the model with a 16384-context length using an instruction dataset, aiming to improve its understanding of QA sessions, and use causal language model approach mentioned at ~\ref{sec:4096-pretrained}. Below are the hyperparameters we use during the fine-tuning process:

\begin{table}[h]
  \centering
  \begin{tabular}{lccl}
    \hline
    \textbf{Hyperparameter} & \textbf{Value} \\
    \hline
    DeepSpeed               & ZeRO-3 Offload \\
    Batch Size              & 6              \\
    Learning Rate           & constant 2e-5  \\

    \hline
  \end{tabular}
\end{table}

Complete fine-tuning 16384 context length implementation at \href{https://github.com/mesolitica/malaya/tree/5.1/session/mistral#instructions-7b-16384-context-length}{here}.

\subsection{Result Finetuning}\label{sec:result-finetuning}

We obtained our results using the fine-tuned model accessible at \href{https://huggingface.co/mesolitica/malaysian-mistral-7b-32k-instructions}{mesolitica/malaysian-mistral-7b-32k-instructions} using text generation parameters below:

\begin{table}[h]
  \centering
  \begin{tabular}{lccl}
    \hline
    \textbf{Parameter} & \textbf{Value} \\
    \hline
    top\_p             & 0.95           \\
    top\_k             & 50             \\
    do\_sample         & True           \\
    temperature        & 0.9            \\
    num\_beams         & 1              \\
    \hline
  \end{tabular}
\end{table}

\subsubsection{Multiturn Malaysian context QA}

\begin{lstlisting}[breaklines=true]
  User: kwsp tu apa
  Assistant: KWSP merujuk kepada Kumpulan Wang Simpanan Pekerja (KWSP) yang ditubuhkan di bawah Akta Kumpulan Wang Simpanan Pekerja 1991. Ia adalah satu skim simpanan untuk pekerja dan majikan di Malaysia yang membantu menyediakan dana persaraan bagi individu pada masa hadapan.
  User: ye ke ko ni
  Assistant: Ya, betul. Saya seorang pembantu pintar yang sedia membantu anda dengan soalan atau permintaan lain yang mungkin anda miliki. Jika anda mempunyai sebarang pertanyaan lagi, jangan teragak-agak untuk bertanya!
  User: salah jawapan ko
  Assistant: Maaf atas kesilapan dalam jawapan saya sebelum ini. Saya akan cuba memberikan jawapan yang lebih tepat dan sesuai. Apakah soalan atau topik yang ingin anda bincangkan? Saya sedia membantu!
\end{lstlisting}

\subsubsection{Multiturn Coding QA}

\begin{lstlisting}[breaklines=true]
  User: camne nak code sentiment analysis guna python
  Assistant: Untuk melaksanakan analisis sentimen menggunakan Python, anda boleh mengikuti langkah-langkah berikut:

  Pemasangan dan Penyediaan Data:
  
  Pasang perpustakaan yang diperlukan seperti NLTK (Natural Language Toolkit) untuk pemprosesan bahasa semula jadi dan scikit-learn untuk model pembelajaran mesin.
  Dapatkan set data teks yang akan digunakan untuk analisis sentimen. Anda boleh mendapatkannya dari sumber awam atau membuatnya sendiri.
  Pra-Pemprosesan Teks:
  
  Bersihkan dan pra-proses teks dengan menghilangkan tanda baca, nombor, dan elemen bukan teks lain.
  Lakukan tokenisasi untuk memecahkan teks ke dalam unit individu (seperti perkataan).
  Buang kata henti (kata umum yang tidak memberikan makna penting seperti "dan", "di", "ke" dll.).
  Gunakan stemming atau lemmatization untuk mengubah perkataan ke bentuk asas mereka.
  Ekstraksi Ciri:
  
  Gunakan kaedah pengekstrakan ciri seperti Bag of Words (BoW), TF-IDF, atau word embeddings (contohnya Word2Vec atau GloVe) untuk mengubah teks menjadi vektor ciri numerik.
  Model Latihan:
  
  Bahagikan set data anda kepada subset latihan dan ujian.
  Latih model pembelajaran mesin menggunakan algoritma seperti Naive Bayes, SVM, atau Neural Networks pada ciri yang telah diekstrak.
  Nilai prestasi model menggunakan metrik seperti ketepatan, skor F1, atau matriks kekeliruan.
  Analisis Sentimen:
  
  Ramal label sentimen untuk set data ujian menggunakan model yang telah dilatih.
  Kira nilai purata sentimen untuk setiap label (positif, negatif, neutral) untuk mendapatkan gambaran keseluruhan tentang sentimen.
  Visualisasi Hasil:
  
  Gunakan perpustakaan visualisasi seperti matplotlib atau seaborn untuk mewakili hasil analisis sentimen secara grafik.
  Berikut adalah contoh ringkas bagaimana anda boleh melaksanakan analisis sentimen menggunakan NLTK dan scikit-learn:
  
  ```
  import numpy as np
  from nltk.tokenize import word_tokenize
  from sklearn.feature_extraction.text import CountVectorizer
  from sklearn.model_selection import train_test_split
  from sklearn.ensemble import RandomForestClassifier
  from sklearn.metrics import classification_report
  
  # Contoh data mudah
  reviews = ["Saya sangat suka telefon ini! Telefon ini sangat hebat.", 
                "Telefon ini agak baik tetapi tidak terbaik."]
  labels = ['positif', 'positif']
  
  # Tokenisasi dan pra-pemprosesan
  tokens = []
  for review in reviews:
      tokens += word_tokenize(review.lower())
  
  # Mewujudkan beg perkataan
  vectorizer = CountVectorizer()
  X = vectorizer.fit_transform(tokens).toarray()
  
  # Membahagi set data kepada set kereta api dan set ujian
  X_train, X_test, y_train, y_test = train_test_split(X, labels, test_size=0.2, random_state=42)
  
  # Melatih model
  model = RandomForestClassifier()
  ``` ... truncated
  User: ada contoh tensorflow untuk sentiment classification tak?
  Assistant: Ya, saya boleh memberikan contoh kod TensorFlow untuk klasifikasi sentimen. Berikut adalah contoh ringkas untuk melatih model rangkaian saraf tiruan (neural network) menggunakan TensorFlow untuk tugas klasifikasi sentimen:
  
  ```
  import tensorflow as tf
  from tensorflow.keras.preprocessing.text import Tokenizer
  from tensorflow.keras.preprocessing.sequence import pad_sequences
  from tensorflow.keras.models import Sequential
  from tensorflow.keras.layers import Embedding, LSTM, Dense
  
  # Contoh data mudah
  reviews = ["Saya sangat suka telefon ini! Telefon ini sangat hebat.",
              "Telefon ini agak baik tetapi tidak terbaik."]
  
  # Label sentimen
  labels = ['positif', 'positif']
  
  # Inisialisasi Tokenizer
  tokenizer = Tokenizer(num_words=1000, oov_token="<OOV>")
  tokenizer.fit_on_texts(reviews)
  word_index = tokenizer.word_index
  
  # Menukar ulasan kepada urutan token
  sequences = tokenizer.texts_to_sequences(reviews)
  padded_sequences = pad_sequences(sequences, maxlen=max_length)
  
  # Membina model neural network
  model = Sequential([
      Embedding(input_dim=1000, output_dim=16, input_length=max_length),
      LSTM(64),
      Dense(1, activation='sigmoid')
  ])
  
  # Kompilasi model
  model.compile(optimizer='adam', loss='binary_crossentropy', metrics=['accuracy'])
  
  # Latih model
  model.fit(padded_sequences, labels, epochs=10, validation_split=0.2)
  
  # Penilaian model
  scores = model.evaluate(padded_sequences, labels)
  print("Accuracy:", scores[1]*100)
  ```

  Dalam contoh ini, kami menggunakan model LSTM (Long Short-Term Memory) untuk mengklasifikasikan sentimen positif atau negatif berdasarkan ulasan. Anda juga boleh menyesuaikan seni bina model dan hiperparameter lain mengikut keperluan anda. Pastikan anda telah memasang TensorFlow sebelum menjalankan kod ini.
\end{lstlisting}

\subsubsection{Malay instruction}

\begin{lstlisting}[breaklines=true]
  User: tolong buat sajak pasal nasi ayam
  Assistant: Nasi ayam, hidangan yang lazat
  Dengan rasa yang menggoda selera
  Ayam panggang, lembut dan beraroma
  Sajian yang tidak boleh ditandingi
  
  Nasi putih dihidangkan dengan nikmat
  Rasanya sempurna dalam setiap suapan
  Tambah kuah ayam yang pekat
  Menyatu dengan nasi, membawa kepuasan
  
  Aroma harum menyeruak dari dapur
  Membuat hati tergerak untuk merasai
  Nasi ayam, hidangan yang istimewa
  Menghilangkan lapar, memenuhi perut yang kosong
\end{lstlisting}

\section{Evaluation}

To assess the performance of our language model, we conducted evaluations on a tatabahasa test set sourced from \url{https://tatabahasabm.tripod.com/latih/latih.htm}, published at \href{https://github.com/mesolitica/malaysian-dataset/tree/master/llm-benchmark/tatabahasabm.tripod.com}{here}. This test set comprises a total of 349 questions, providing a substantial and varied benchmark for evaluating the model's proficiency in handling Tatabahasa (Malay grammar) in the Malay language. An example of the test set,

\begin{lstlisting}[breaklines=true]
{'question': '........, sudah dapat memandu kereta rupa-rupanya kamu !',
 'instruction': None,
 'choices': {'A': {'text': 'Oh', 'answer': False},
  'B': {'text': 'Eh', 'answer': True},
  'C': {'text': 'Hai', 'answer': False},
  'D': {'text': 'Ah', 'answer': False}},
 'website': 'https://tatabahasabm.tripod.com/latih/kseruc.htm'}
\end{lstlisting}

We use the same finetuned model and text generation parameters from section ~\ref{sec:result-finetuning}, \textbf{and each question we generated for 5 times and pick the best answer based on most voted answer}.

\begin{table}[h]
  \centering
  \begin{tabular}{lccl}
    \hline
    \textbf{Model}       & \textbf{Tatabahasa 0 shot} & \textbf{Tatabahasa 1 shot} & \textbf{Tatabahasa 3 shots} \\
    \hline
    gpt-4-1106-preview   & 75.645                     & 73.638                     & 75.644                      \\
    gpt-3.5-turbo-0613   & 59.531                     & 60.806                     & 63.037                      \\
    AWS Bedrock Claude 2 & 61.702                     & 60.171                     & 59.598                      \\
    Malaysian Mistral    & \textbf{65.33}             & 57.306                     & 56.446                      \\
    \hline
  \end{tabular}
\end{table}

We also compared with other models and published the benchmark at \href{https://huggingface.co/spaces/mesolitica/malay-llm-leaderboard}{mesolitica/malay-llm-leaderboard}.

\section{Acknowledgement}

Special thanks to Malaysia-AI volunteers especially \href{https://www.linkedin.com/in/wan-adzhar-faiq-adzlan-19a27baa/}{Wan Adzhar Faiq Adzlan}, \href{https://www.linkedin.com/in/ammar-azman/}{Ammar Azman}, \href{https://www.linkedin.com/in/amzar96/}{M. Amzar}, \href{https://www.linkedin.com/in/muhammad-farhan-helmy-0529501a7/}{Muhammad Farhan} and \href{https://www.linkedin.com/in/syafie-nizam/}{Syafie Nizam} for contributing dataset to train Malaysian Mistral.

We would like to express our gratitude to NVIDIA Inception for generously providing us with the opportunity to train our model on the Azure cloud. Their support has played a crucial role in the success of our research, enabling us to leverage advanced technologies and computational resources.

We extend our thanks to the wider research community for their valuable insights and collaborative discussions, which have greatly influenced our work. This paper reflects the collective efforts and contributions from both NVIDIA Inception and the broader research community.

\section{Conclusion}

In summary, creating a large language model tailored for Malaysia, was a detailed process involving pretraining and finetuning on a specific dataset. This effort required significant time, funding, and expertise. Training Mistral demanded high-end hardware, such as 8 A100 GPUs, making it a State of the Art model surpassing ChatGPT3.5 in the Malaysian context. The decision to open-source Malaysian Mistral and its code reflects a commitment to accessibility and collaboration, allowing both public and private organizations to benefit. Looking forward, the team plans to introduce an open-sourced multi-modal model capable of processing audio, image, and text. Positioned at the forefront of AI in Malaysia, the goal is to provide cutting-edge models for the public's benefit, showcasing Malaysian Mistral as a technological achievement and a commitment to advancing AI for all.

\bibliography{neurips_2023}{}

\begin{thebibliography}{1}

\bibitem{vaswani2023attention}
Ashish Vaswani, Noam Shazeer, Niki Parmar, Jakob Uszkoreit, Llion Jones,
  Aidan~N. Gomez, Lukasz Kaiser, and Illia Polosukhin.
\newblock Attention is all you need, 2023.

\bibitem{jiang2023mistral}
Albert~Q. Jiang, Alexandre Sablayrolles, Arthur Mensch, Chris Bamford,
  Devendra~Singh Chaplot, Diego de~las Casas, Florian Bressand, Gianna Lengyel,
  Guillaume Lample, Lucile Saulnier, Lélio~Renard Lavaud, Marie-Anne Lachaux,
  Pierre Stock, Teven~Le Scao, Thibaut Lavril, Thomas Wang, Timothée Lacroix,
  and William~El Sayed.
\newblock Mistral 7b, 2023.

\bibitem{Malaya}
Zolkepli Husein.
\newblock Malaya, natural-language-toolkit library for bahasa malaysia, powered
  by pytorch.
\newblock \url{https://github.com/huseinzol05/malaya}, 2018.

\bibitem{luo2023wizardcoder}
Ziyang Luo, Can Xu, Pu~Zhao, Qingfeng Sun, Xiubo Geng, Wenxiang Hu, Chongyang
  Tao, Jing Ma, Qingwei Lin, and Daxin Jiang.
\newblock Wizardcoder: Empowering code large language models with
  evol-instruct, 2023.

\bibitem{talmor2019commonsenseqa}
Alon Talmor, Jonathan Herzig, Nicholas Lourie, and Jonathan Berant.
\newblock Commonsenseqa: A question answering challenge targeting commonsense
  knowledge, 2019.

\bibitem{xu2023wizardlm}
Can Xu, Qingfeng Sun, Kai Zheng, Xiubo Geng, Pu~Zhao, Jiazhan Feng, Chongyang
  Tao, and Daxin Jiang.
\newblock Wizardlm: Empowering large language models to follow complex
  instructions, 2023.

\bibitem{ding2023enhancing}
Ning Ding, Yulin Chen, Bokai Xu, Yujia Qin, Zhi Zheng, Shengding Hu, Zhiyuan
  Liu, Maosong Sun, and Bowen Zhou.
\newblock Enhancing chat language models by scaling high-quality instructional
  conversations, 2023.

\end{thebibliography}
\bibliographystyle{unsrt}

\end{document}